\documentclass{article}

\PassOptionsToPackage{numbers, compress}{natbib}

\usepackage[final]{styles/neurips_2024}

\usepackage{amsmath}
\usepackage{multirow}
\usepackage{graphicx}
\usepackage{xcolor}
\usepackage{booktabs} 
\usepackage[colorlinks=true, allcolors=blue]{hyperref}

\usepackage{listings}
\title{Docling Technical Report}

\begin{document}

\maketitle

\begin{figure}
\centering
\includegraphics[width=0.25\linewidth]{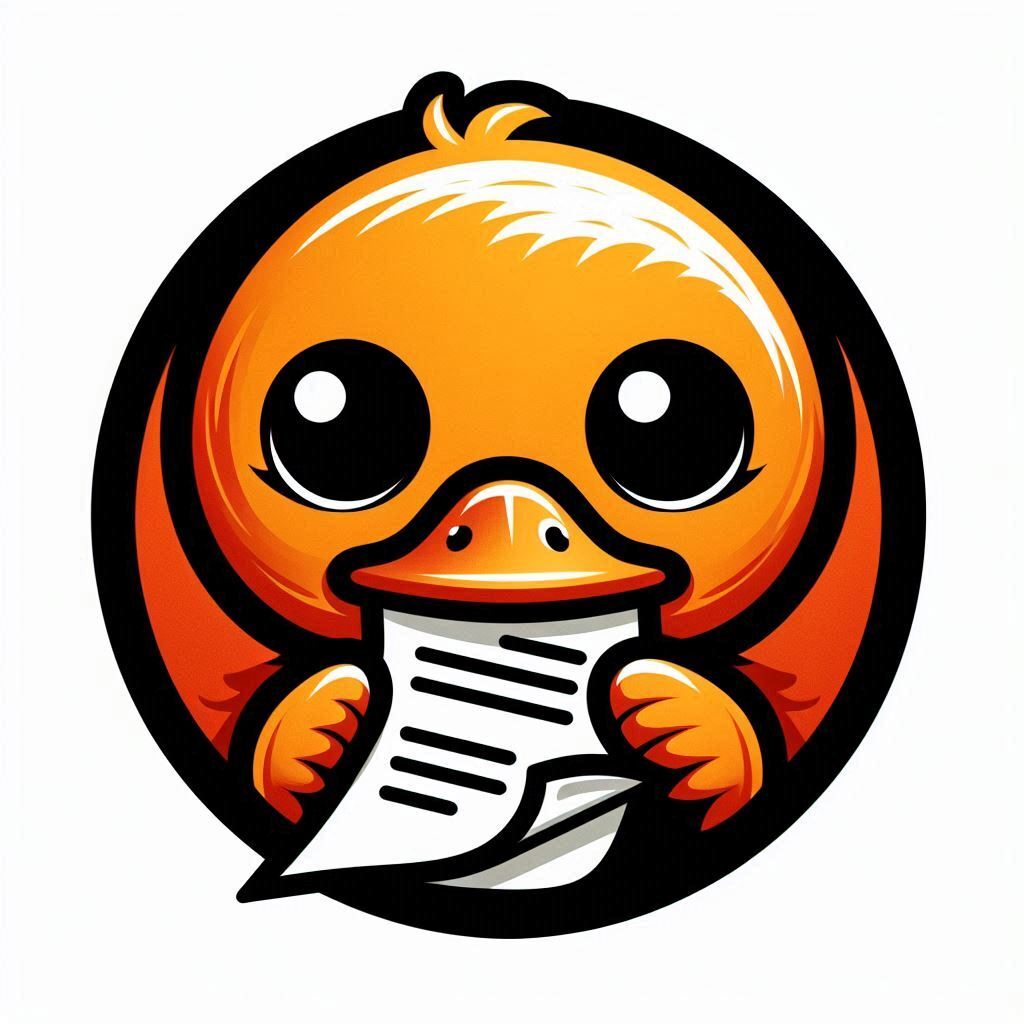}
\end{figure}

\vspace{-2cm}
\begin{center}
Version 1.0 \\
\vspace{2em}
\textbf{Christoph~Auer}\quad
\textbf{Maksym~Lysak}\quad
\textbf{Ahmed~Nassar}\quad
\textbf{Michele~Dolfi}\quad
\textbf{Nikolaos~Livathinos}\quad
\textbf{Panos~Vagenas}\quad
\textbf{Cesar~Berrospi~Ramis}\quad
\textbf{Matteo~Omenetti}\quad
\textbf{Fabian~Lindlbauer}\quad
\textbf{Kasper~Dinkla}\quad
\textbf{Lokesh~Mishra}\quad
\textbf{Yusik~Kim}\quad
\textbf{Shubham~Gupta}\quad
\textbf{Rafael~Teixeira~de~Lima}\quad
\textbf{Valery~Weber}\quad
\textbf{Lucas~Morin}\quad
\textbf{Ingmar~Meijer}\quad
\textbf{Viktor~Kuropiatnyk}\quad
\textbf{Peter~W.~J.~Staar}\quad \\
\vspace{1em}
AI4K Group, IBM Research\\
Rüschlikon, Switzerland \\
\end{center}

\begin{abstract}
This technical report introduces \textit{Docling}, an easy to use, self-contained, MIT-licensed open-source package for PDF document conversion. It is powered by state-of-the-art specialized AI models for layout analysis (DocLayNet) and table structure recognition (TableFormer), and runs efficiently on commodity hardware in a small resource budget. The code interface allows for easy extensibility and addition of new features and models. 

\end{abstract}

\section{Introduction}

Converting PDF documents back into a machine-processable format has been a major challenge for decades due to their huge variability in formats, weak standardization and printing-optimized characteristic, which discards most structural features and metadata. With the advent of LLMs and popular application patterns such as retrieval-augmented generation (RAG), leveraging the rich content embedded in PDFs has become ever more relevant.
In the past decade, several powerful document understanding solutions have emerged on the market, most of which are commercial software, cloud offerings~~\cite{auer2022delivering} and most recently, multi-modal vision-language models. As of today, only a handful of open-source tools cover PDF conversion, leaving a significant feature and quality gap to proprietary solutions. 

With \textit{Docling}, we open-source a very capable and efficient document conversion tool which builds on the powerful, specialized AI models and datasets for layout analysis and table structure recognition we developed and presented in the recent past~~\cite{tableformer, dln, lysak_optimized_2023}. Docling is designed as a simple, self-contained python library with permissive license, running entirely locally on commodity hardware. Its code architecture allows for easy extensibility and addition of new features and models. 

\pagebreak
Here is what Docling delivers today:

\begin{itemize}
    \item Converts PDF documents to JSON or Markdown format, stable and lightning fast
    \item Understands detailed page layout, reading order, locates figures and recovers table structures
    \item Extracts metadata from the document, such as title, authors, references and language
    \item Optionally applies OCR, e.g. for scanned PDFs
    \item Can be configured to be optimal for batch-mode (i.e high throughput, low time-to-solution) or interactive mode (compromise on efficiency, low time-to-solution)
    \item Can leverage different accelerators (GPU, MPS, etc).
\end{itemize}

\section{Getting Started}

To use Docling, you can simply install the \textit{\href{https://pypi.org/project/docling/}{docling}} package from PyPI. Documentation and examples are available in our \href{https://github.ibm.com/DS4SD/docling}{GitHub} repository at \href{https://github.com/DS4SD/docling}{github.com/DS4SD/docling}. All required model assets\footnote{see \href{https://huggingface.co/ds4sd/docling-models/}{huggingface.co/ds4sd/docling-models/}} are downloaded to a local huggingface datasets cache on first use, unless you choose to pre-install the model assets in advance.

Docling provides an easy code interface to convert PDF documents from file system, URLs or binary streams, and retrieve the output in either JSON or Markdown format. For convenience, separate methods are offered to convert single documents or batches of documents. A basic usage example is illustrated below. Further examples are available in the Doclign code repository.

\begin{lstlisting}[language=Python]
from docling.document_converter import DocumentConverter

source = "https://arxiv.org/pdf/2206.01062"  # PDF path or URL
converter = DocumentConverter()
result = converter.convert_single(source)
print(result.render_as_markdown())  # output: "## DocLayNet: A Large Human-Annotated Dataset for Document-Layout Analysis [...]"
\end{lstlisting}

Optionally, you can configure custom pipeline features and runtime options, such as turning on or off features (e.g. OCR, table structure recognition), enforcing limits on the input document size, and defining the budget of CPU threads. Advanced usage examples and options are documented in the README file. Docling also provides a \textit{Dockerfile} to demonstrate how to install and run it inside a container.

\section{Processing pipeline}

\begin{figure}
    \centering
    \includegraphics[width=1\linewidth]{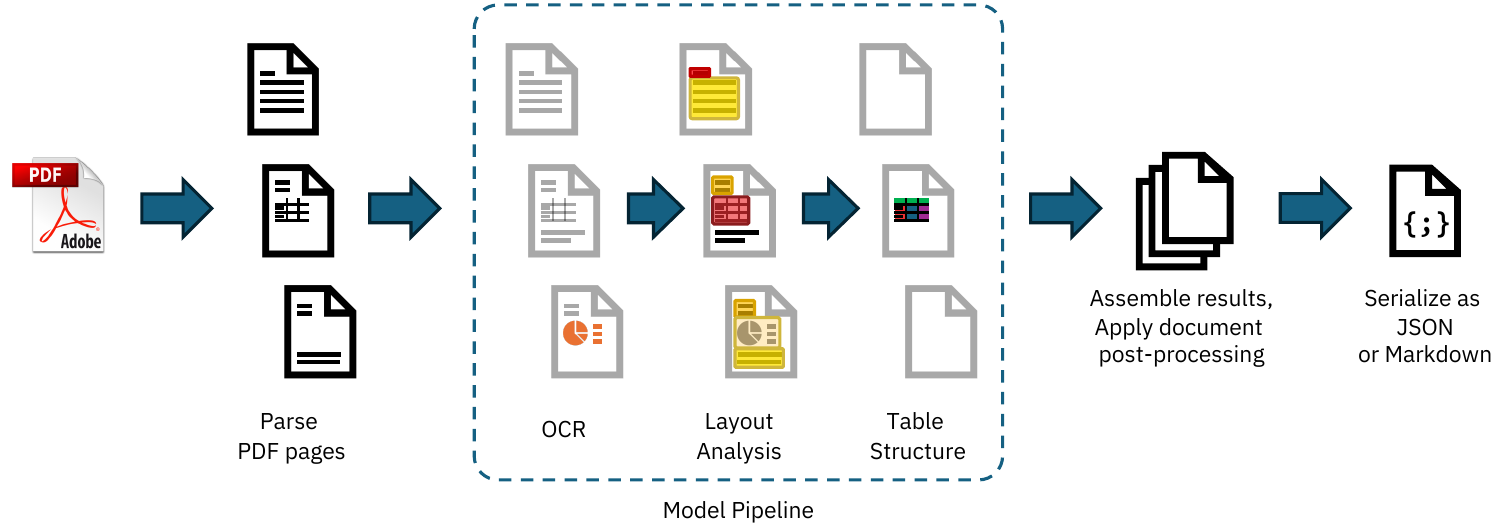}
    \caption{Sketch  of Docling's default processing pipeline. The inner part of the model pipeline is easily customizable and extensible.}
    \label{fig:pipeline}
\end{figure}

Docling implements a linear pipeline of operations, which execute sequentially on each given document (see Fig.~\ref{fig:pipeline}). Each document is first parsed by a PDF backend, which retrieves the programmatic text tokens, consisting of string content and its coordinates on the page, and also renders a bitmap image of each page to support downstream operations.
Then, the standard model pipeline applies a sequence of AI models independently on every page in the document to extract features and content, such as layout and table structures. Finally, the results from all pages are aggregated and passed through a post-processing stage, which augments metadata, detects the document language, infers reading-order and eventually assembles a typed document object which can be serialized to JSON or Markdown.

\subsection{PDF backends}

Two basic requirements to process PDF documents in our pipeline are a) to retrieve all text content and their geometric coordinates on each page and b) to render the visual representation of each page as it would appear in a PDF viewer. Both these requirements are encapsulated in Docling's PDF backend interface. While there are several open-source PDF parsing libraries available for python, we faced major obstacles with all of them for different reasons, among which were restrictive licensing (e.g. pymupdf~\cite{pymupdf}), poor speed or unrecoverable quality issues, such as merged text cells across far-apart text tokens or table columns (pypdfium, PyPDF)~\cite{pypdfium2,pypdf}.

We therefore decided to provide multiple backend choices, and additionally open-source a custom-built PDF parser, which is based on the low-level \textit{qpdf}\cite{qpdf} library. It is made available in a separate package named \textit{\href{https://pypi.org/project/docling-parse/}{docling-parse}} and powers the default PDF backend in Docling. As an alternative, we provide a PDF backend relying on \textit{pypdfium}, which may be a safe backup choice in certain cases, e.g. if issues are seen with particular font encodings.

\subsection{AI models}

As part of Docling, we initially release two highly capable AI models to the open-source community, which have been developed and published recently by our team. The first model is a layout analysis model, an accurate object-detector for page elements~\cite{dln}. The second model is TableFormer~\cite{tableformer,lysak_optimized_2023}, a state-of-the-art table structure recognition model. We provide the pre-trained weights (hosted on \href{https://huggingface.co/ds4sd/docling-models/}{huggingface}) and a separate package for the inference code as \textit{\href{https://github.com/DS4SD/docling-ibm-models}{docling-ibm-models}}. Both models are also powering the open-access \href{https://ds4sd.github.io}{deepsearch-experience}, our cloud-native service for knowledge exploration tasks. 

\subsubsection*{Layout Analysis Model}
Our layout analysis model is an object-detector which predicts the bounding-boxes and classes of various elements on the image of a given page. Its architecture is derived from RT-DETR~\cite{lv2023detrs} and re-trained on DocLayNet~\cite{dln}, our popular human-annotated dataset for document-layout analysis, among other proprietary datasets. For inference, our implementation relies on the \textit{onnxruntime}~\cite{onnxruntime}.

The Docling pipeline feeds page images at 72 dpi resolution, which can be processed on a single CPU with sub-second latency. All predicted bounding-box proposals for document elements are post-processed to remove overlapping proposals based on confidence and size, and then intersected with the text tokens in the PDF to group them into meaningful and complete units such as paragraphs, section titles, list items, captions, figures or tables.

\subsubsection*{Table Structure Recognition}
The TableFormer model~\cite{tableformer}, first published in 2022 and since refined with a custom structure token language~\cite{lysak_optimized_2023}, is a vision-transformer model for table structure recovery. It can predict the logical row and column structure of a given table based on an input image, and determine which table cells belong to column headers, row headers or the table body. Compared to earlier approaches, TableFormer handles many characteristics of tables, such as partial or no borderlines, empty cells, rows or columns, cell spans and hierarchy both on column-heading or row-heading level, tables with inconsistent indentation or alignment and other complexities. For inference, our implementation relies on \textit{PyTorch}~\cite{pytorch2}.

The Docling pipeline feeds all table objects detected in the layout analysis to the TableFormer model, by providing an image-crop of the table and the included text cells. TableFormer structure predictions are matched back to the PDF cells in post-processing to avoid expensive re-transcription text in the table image. Typical tables require between 2 and 6 seconds to be processed on a standard CPU, strongly depending on the amount of included table cells.


\subsubsection*{OCR}

Docling provides optional support for OCR, for example to cover scanned PDFs or content in bitmaps images embedded on a page. In our initial release, we rely on \textit{EasyOCR}~\cite{easyocr}, a popular third-party OCR library with support for many languages. Docling, by default, feeds a high-resolution page image (216 dpi) to the OCR engine, to allow capturing small print detail in decent quality. While EasyOCR delivers reasonable transcription quality, we observe that it runs fairly slow on CPU (upwards of 30 seconds per page). 

We are actively seeking collaboration from the open-source community to extend Docling with additional OCR backends and speed improvements.

\subsection{Assembly}

In the final pipeline stage, Docling assembles all prediction results produced on each page into a well-defined datatype that encapsulates a converted document, as defined in the auxiliary package \textit{\href{https://github.com/DS4SD/docling-core}{docling-core}}. The generated document object is passed through a post-processing model which leverages several algorithms to augment features, such as detection of the document language, correcting the reading order, matching figures with captions and labelling metadata such as title, authors and references.
The final output can then be serialized to JSON or transformed into a Markdown representation at the users request.

\subsection{Extensibility}

Docling provides a straight-forward interface to extend its capabilities, namely the model pipeline. A model pipeline constitutes the central part in the processing, following initial document parsing and preceding output assembly, and can be fully customized by sub-classing from an abstract baseclass (\textit{BaseModelPipeline}) or cloning the default model pipeline. This effectively allows to fully customize the chain of models, add or replace models, and introduce additional pipeline configuration parameters. To use a custom model pipeline, the custom pipeline class to instantiate can be provided as an argument to the main document conversion methods. We invite everyone in the community to propose additional or alternative models and improvements.

Implementations of model classes must satisfy the python \lstinline{Callable} interface. The \lstinline{__call__} method must accept an iterator over page objects, and produce another iterator over the page objects which were augmented with the additional features predicted by the model, by extending the provided \lstinline{PagePredictions} data model accordingly. 


\section{Performance}

In this section, we establish some reference numbers for the processing speed of Docling and the resource budget it requires. All tests in this section are run with default options on our standard test set distributed with Docling, which consists of three papers from arXiv and two IBM Redbooks, with a total of 225 pages. Measurements were taken using both available PDF backends on two different hardware systems: one MacBook Pro M3 Max, and one bare-metal server running Ubuntu 20.04 LTS on an Intel Xeon E5-2690 CPU. For reproducibility, we fixed the thread budget (through setting \textit{OMP\_NUM\_THREADS environment variable}) once to 4 (Docling default) and once to 16 (equal to full core count on the test hardware). All results are shown in Table~\ref{tab:performance}.

If you need to run Docling in very low-resource environments, please consider configuring the pypdfium backend. While it is faster and more memory efficient than the default \textit{docling-parse} backend, it will come at the expense of worse quality results, especially in table structure recovery.

Establishing GPU acceleration support for the AI models is currently work-in-progress and largely untested, but may work implicitly when CUDA is available and discovered by the onnxruntime and torch runtimes backing the Docling pipeline. We will deliver updates on this topic at in a future version of this report.


\begin{table}[htbp]
\centering
\caption{Runtime characteristics of Docling with the standard model pipeline and settings, on our test dataset of 225 pages, on two different systems. OCR is disabled. We show the time-to-solution (TTS), computed throughput in pages per second, and the peak memory used (resident set size) for both the Docling-native PDF backend and for the pypdfium backend, using 4 and 16 threads.}
\label{tab:performance}
\begin{tabular}{@{}llcccccc@{}}
\toprule
\multirow{2}{*}{CPU} & \multirow{2}{*}{\begin{tabular}[c]{@{}l@{}}Thread\\budget\end{tabular}} & \multicolumn{3}{c}{native backend} & \multicolumn{3}{c}{pypdfium backend} \\
\cmidrule(lr){3-5} \cmidrule(l){6-8}
 &  & TTS & Pages/s & Mem & TTS & Pages/s & Mem \\
\midrule
\multirow{2}{*}{\begin{tabular}[c]{@{}l@{}}Apple M3 Max\\(16 cores)\end{tabular}} 
 & 4  & 177 s & 1.27 & \multirow{2}{*}{6.20 GB} & 103 s & 2.18 & \multirow{2}{*}{2.56 GB} \\
 & 16 & 167 s & 1.34 &                          & 92 s  & 2.45 &                          \\
\midrule
\multirow{3}{*}{\begin{tabular}[c]{@{}l@{}}Intel(R) Xeon\\E5-2690\\(16 cores)\end{tabular}} 
 & 4  & 375 s & 0.60 & \multirow{2}{*}{6.16 GB} & 239 s & 0.94 & \multirow{2}{*}{2.42 GB} \\
 & 16 & 244 s & 0.92 &                          & 143 s & 1.57 &                          \\
\end{tabular}
\end{table}

\section{Applications}

Thanks to the high-quality, richly structured document conversion achieved by Docling, its output qualifies for numerous downstream applications. For example, Docling can provide a base for detailed enterprise document search, passage retrieval or classification use-cases, or support knowledge extraction pipelines, allowing specific treatment of different structures in the document, such as tables, figures, section structure or references. 
For popular generative AI application patterns, such as retrieval-augmented generation (RAG), we provide \textit{\href{https://pypi.org/project/quackling/}{quackling}}, an open-source package which capitalizes on Docling's feature-rich document output to enable document-native optimized vector embedding and chunking. It plugs in seamlessly with LLM frameworks such as LlamaIndex~\cite{Liu_LlamaIndex_2022}.
Since Docling is fast, stable and cheap to run, it also makes for an excellent choice to build document-derived datasets. With its powerful table structure recognition, it provides significant benefit to automated knowledge-base construction~\cite{Morin2024,mishra-etal-2024-statements}.   
Docling is also integrated within the open IBM data prep kit~\cite{dpk}, which implements scalable data transforms to build large-scale multi-modal training datasets.

\section{Future work and contributions}

Docling is designed to allow easy extension of the model library and pipelines. In the future, we plan to extend Docling with several more models, such as a figure-classifier model, an equation-recognition model, a code-recognition model and more. This will help improve the quality of conversion for specific types of content, as well as augment extracted document metadata with additional information. Further investment into testing and optimizing GPU acceleration as well as improving the Docling-native PDF backend are on our roadmap, too.

\textbf{We encourage everyone to propose or implement additional features and models, and will gladly take your inputs and contributions under review}. The codebase of Docling is open for use and contribution, under the MIT license agreement and in alignment with our contributing guidelines included in the Docling repository. If you use Docling in your projects, please consider citing this technical report.

\bibliographystyle{abbrvnat}
\bibliography{main}

\newpage
\section*{Appendix\label{sec:Appendix}}

In this section, we illustrate a few examples of Docling's output in Markdown and JSON.

\begin{figure}[h!]
    \centering
    \includegraphics[width=1.1\linewidth]{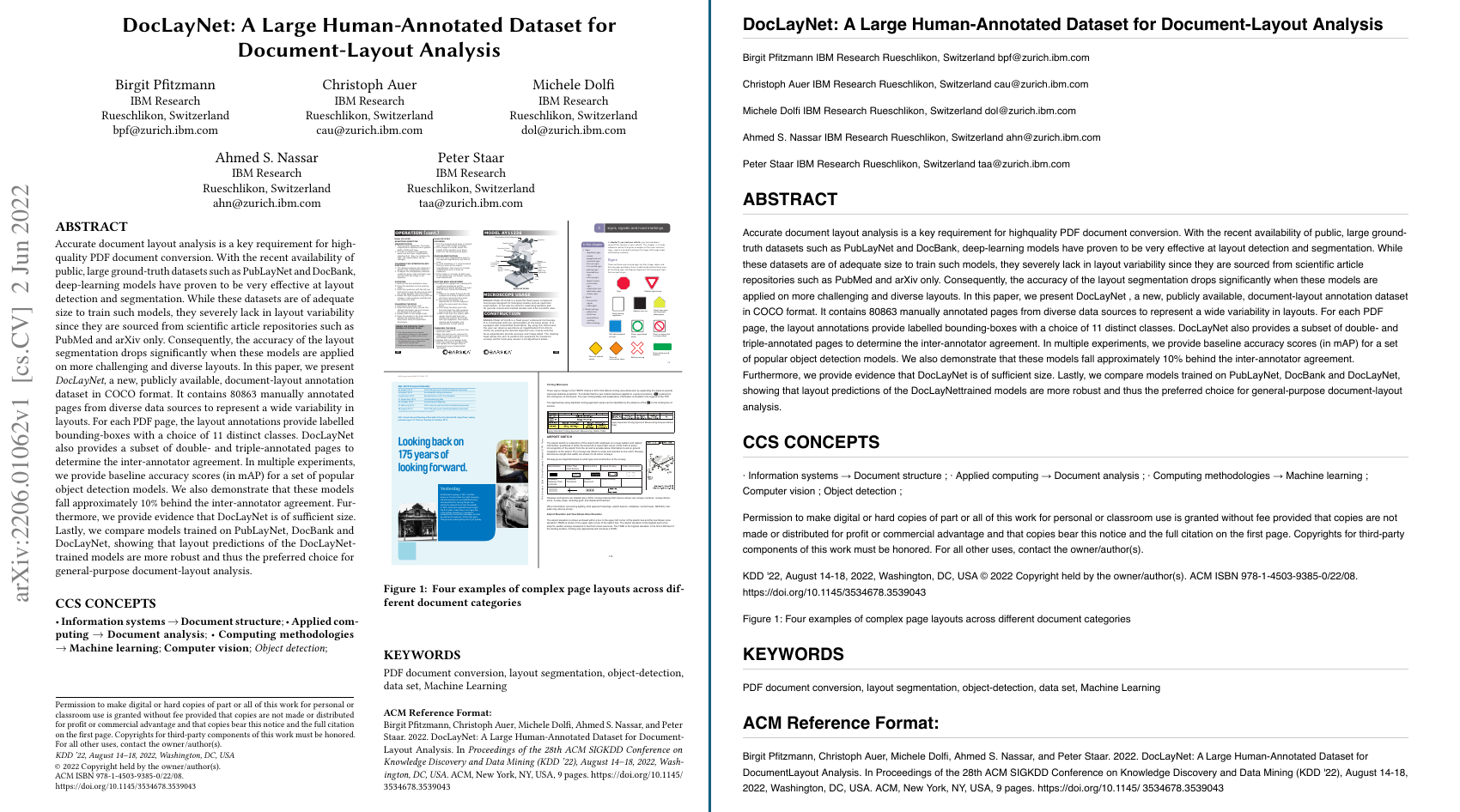}
    \caption{Title page of the DocLayNet paper (\href{https://arxiv.org/pdf/2206.01062}{arxiv.org/pdf/2206.01062}) - left PDF, right rendered Markdown. If recognized, metadata such as authors are appearing first under the title. Text content inside figures is currently dropped, the caption is retained and linked to the figure in the JSON representation (not shown).}
    \label{fig:example1}
\end{figure}

\begin{figure}
    \centering
    \includegraphics[width=1.1\linewidth]{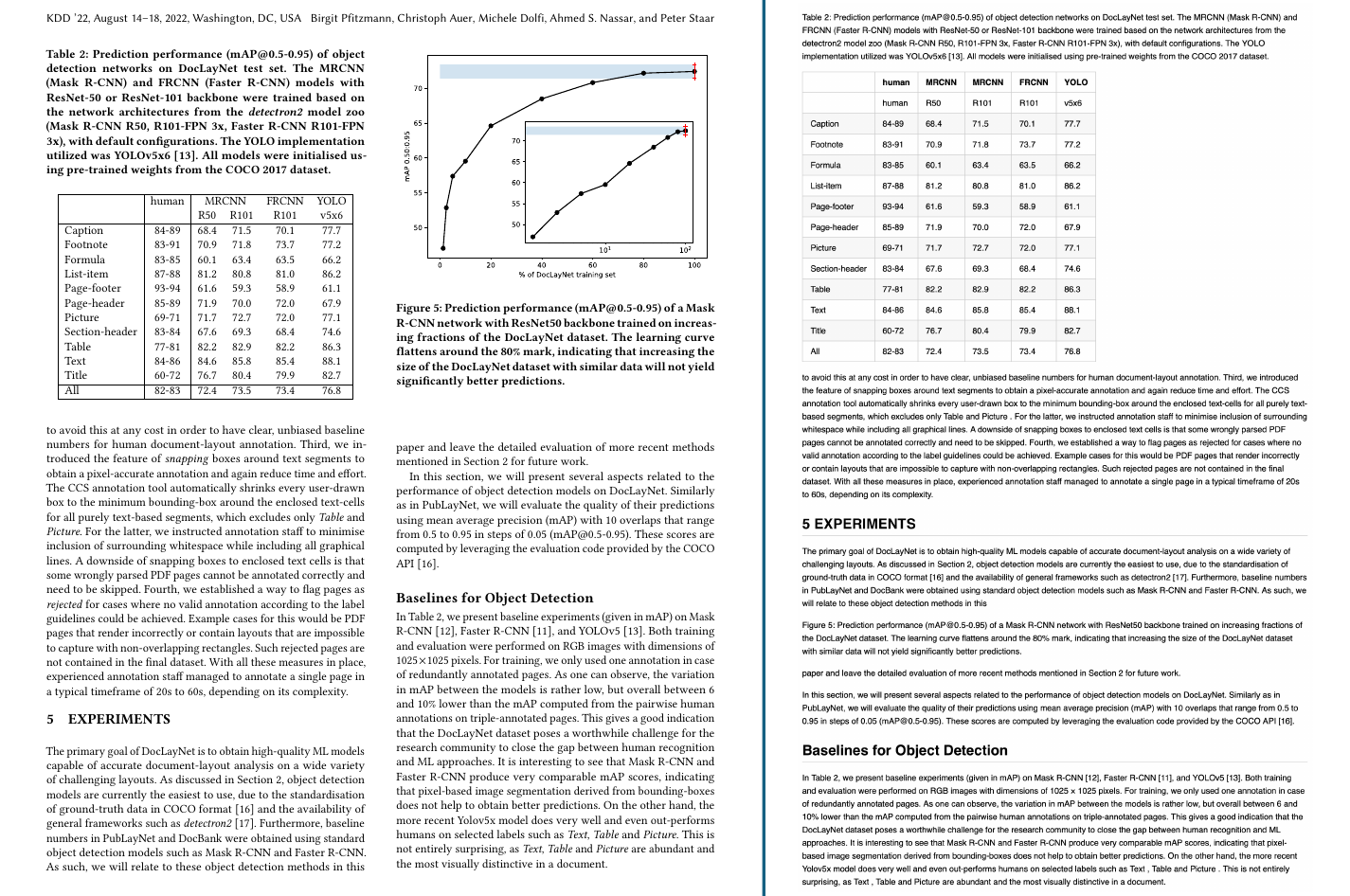}
    \caption{Page 6 of the DocLayNet paper. If recognized, metadata such as authors are appearing first under the title. Elements recognized as page headers or footers are suppressed in Markdown to deliver uninterrupted content in reading order. Tables are inserted in reading order. The paragraph in "5. Experiments" wrapping over the column end is broken up in two and interrupted by the table.}
    \label{fig:example2}
\end{figure}

\begin{figure}
    \centering
    \includegraphics[width=1.1\linewidth]{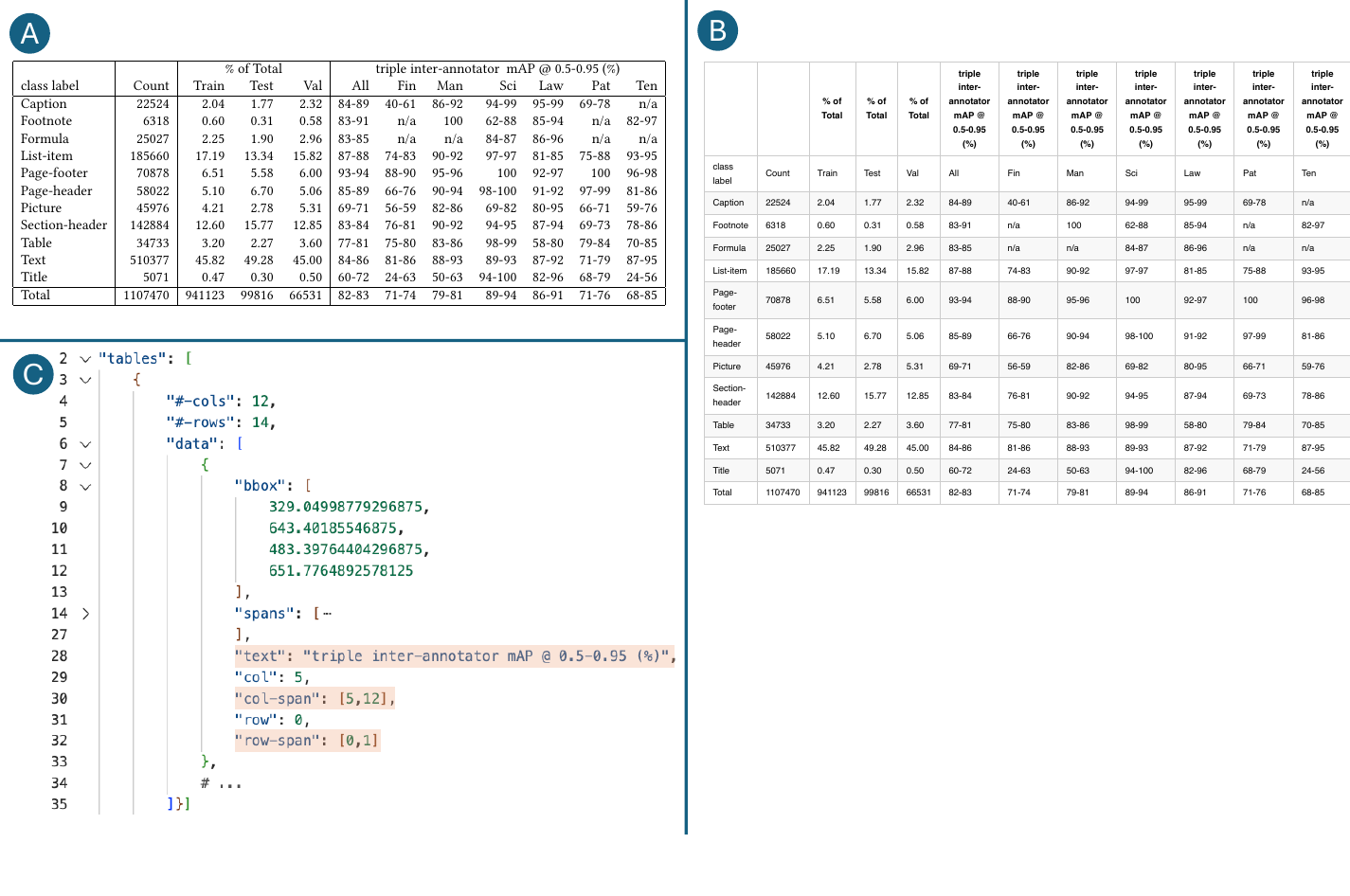}
    \caption{Table 1 from the DocLayNet paper in the original PDF (A), as rendered Markdown (B) and in JSON representation (C). Spanning table cells, such as the multi-column header "triple inter-annotator mAP@0.5-0.95 (\%)", is repeated for each column in the Markdown representation (B), which guarantees that every data point can be traced back to row and column headings only by its grid coordinates in the table. In the JSON representation, the span information is reflected in the fields of each table cell (C). 
    }
    \label{fig:example3}
\end{figure}

\end{document}